\documentclass[a4paper,fleqn]{cas-dc}
\usepackage{listings}
\usepackage{amssymb}
\usepackage{hyperref}
\usepackage{graphicx}
\usepackage{multirow}
\usepackage{booktabs}
\usepackage{xurl}
\usepackage{lipsum}
\usepackage{lscape}
\usepackage{amsmath}
\usepackage{relsize}%
\usepackage[figuresright]{rotating}
\usepackage{lineno}
\usepackage{algorithm} 
\usepackage{algpseudocode} 
\usepackage{comment}
\usepackage{lineno}
\usepackage{nomencl}
\makenomenclature
\usepackage{calc}
\usepackage[T1]{fontenc}
\usepackage{relsize}
\usepackage{subfigure}
\usepackage{tikz}
\usetikzlibrary{fit}
\usepackage{float}

\usepackage{tabularx}
\usepackage{array}

\usepackage{lineno}

\usepackage{natbib}
\setcitestyle{authoryear,open={(},close={)},citesep={;}} 

\hypersetup{
  colorlinks,
  citecolor=Violet,
  linkcolor=Red,
  urlcolor=Blue}

\newcolumntype{P}[1]{>{\centering\arraybackslash}p{#1}}

\def\tsc#1{\csdef{#1}{\textsc{\lowercase{#1}}\xspace}}
\tsc{WGM}
\tsc{QE}
\tsc{EP}
\tsc{PMS}
\tsc{BEC}
\tsc{DE}

\begin{document}

\let\WriteBookmarks\relax
\def\floatpagepagefraction{1}
\def\textpagefraction{.001}
% \shorttitle{}        %  
% \shortauthors{}      %  

\shorttitle{Real-Time Weed Detection in Soybean Fields}   
\shortauthors{Linyuan et~al.}
 
\title [mode = title]{Resource-Constrained UAV-Based Weed Detection for Site-Specific Management on Edge Devices}

\author[1]{Linyuan Wang}\ead{lw1789@msstate.edu}
\author[2]{Haibo Yao}\ead{Haibo.Yao@usda.gov}
\author[3]{Te-Ming Tseng}\ead{tt1024@msstate.edu}
\author[4,5]{Kelvin Betitame}\ead{kelvin.betitame@ndsu.edu} 
\author[4,5]{Xin Sun}\ead{xin.sun@ndsu.edu} 
\author[2]{Yanbo Huang}\ead{yanbo.huang@usda.gov}
\author[1]{Dong Chen\corref{cor1}}\ead{dc2528@msstate.edu} 

\address[1]{Department of Agricultural and Biological Engineering, Mississippi State University, Mississippi State, MS, USA}
\address[2]{USDA-ARS, Genetics and Sustainable Agriculture Research Unit, Mississippi State, MS, USA}
\address[3]{Department of Plant and Soil Sciences, Mississippi State University, Mississippi State, MS, USA}
\address[4]{Department of Agricultural and Biosystems Engineering, North Dakota State University, Fargo, ND, USA}
\address[5]{NDSU Peltier Institute for the Advancement of Agricultural Technology, North Dakota State University, Fargo, ND, USA}

\address{*Dong Chen is the corresponding author}

\begin{abstract}
Weeds compete with crops for light, water, and nutrients, resulting in reduced yield and crop quality. Efficient weed detection is critical for enabling site-specific weed management (SSWM) in crop production. While recent studies have demonstrated the feasibility of deploying deep learning models on edge systems based on Unmanned Aerial Vehicle  (UAV), there remains a critical lack of systematic understanding of how different model architectures behave under real-world resource constraints.
To address this gap, this study presents a deployment-oriented framework for real-time UAV-based weed detection on resource-constrained edge platforms. The framework integrates UAV-based data acquisition, model development, and on-device inference, with an emphasis on balancing detection accuracy and computational efficiency for practical field deployment. A diverse set of state-of-the-art object detection architectures, including convolutional models from the You Only Look Once (YOLO) family (v8–v12) and transformer-based models such as Real-Time Detection Transformer (RT-DETR) (v1–v2), are systematically evaluated to characterize their performance across different edge computing scenarios.
Experiments on three representative edge devices, NVIDIA Jetson Orin Nano, Jetson AGX Xavier, and Jetson AGX Orin, reveal clear trade-offs between accuracy and inference latency across model families and hardware configurations. Results show that high-capacity models achieve up to 86.9\% $\text{mAP}_{50}$, but incur significantly higher latency, limiting their real-time applicability on edge devices. In contrast, lightweight models achieve 66\%–71\% $\text{mAP}_{50}$ while delivering substantially lower inference latency, enabling real-time performance in field conditions. Among the evaluated models, RT-DETRv2-R50-M achieves competitive accuracy ($79\%$ $\text{mAP}_{50}$) with improved efficiency compared to comparable YOLO variants, while lightweight models such as YOLOv10n demonstrate the fastest inference speeds. Notably, YOLOv11s and RT-DETRv2-R50-M provide the most favorable trade-off between accuracy and speed, making them strong candidates for real-time UAV deployment.

\end{abstract}
\maketitle

\section{Introduction}

% Efficient UAV Weed Detection for Site-Specific Management under Edge Constraints
% Resource-Constrained Edge Computing: Benchmarking Object Detectors for UAV-Based Weed Management
% Evaluating Edge Deployment for UAV-Based Weed Detection under Hardware Constraints
% Balancing Performance in Edge Deployment for UAV Weed Detection under Hardware Constraints 
% Edge Deployment for UAV Weed Detection under Hardware Constraints: A Benchmark Study
\paragraph{}
% soybean importance 

% Weeds Pressure
Weed pressure remains one of the major agronomic challenges in crop production, primarily due to competition for light, water, and nutrients. Under uncontrolled conditions, weed interference has been estimated to cause an average potential yield loss of approximately 52\% \citep{soltani2017perspectives}. Herbicide application remains the dominant strategy for weed management in modern cropping systems \citep{ugljic2025stakeholder}. In practice, this often involves broadcast applications across entire fields. Although such applications can suppress weeds, they are poorly matched to the patchy, spatially aggregated distribution of weeds across fields \citep{thorp2004review}. As a result, herbicides may be unnecessarily applied to weed-free areas, potentially disturbing soil microbial communities and nutrient-cycling processes and affecting soil health and long-term crop productivity \citep{ni2025increasing}. Effective herbicide application depends on both the spatial distribution and the specific composition of weed populations. Since herbicide efficacy varies between grassy and broadleaf species \citep{win2023control}, the need to distinguish these functional groups alongside their patchy distribution in the field significantly increases the difficulty of management planning in production.

Machine vision has become essential to precision agriculture, providing the visual perception needed for automated field operations \citep{mavridou2019machine, li2024performance}. Modern agricultural vision systems typically use RGB \citep{ahmad2021performance} or multispectral \citep{candiago2015evaluating} sensors to capture high-resolution field images. By analyzing plant characteristics such as shape, texture, and spectral signature, the systems can distinguish crops from weeds and locate their exact position \citep{wang2019review}. 
As a result, vision-based detection has become a key component in the precision weed control while reducing unnecessary herbicide application \citep{gerhards2022advances}.

Real-world field environments are often characterized by significant illumination variations \citep{sun2024evaluation}, complex soil backgrounds \citep{shuai2025yolo}, and dense crop occlusion \citep{lottes2018fully}. These challenges jointly limit the accuracy and robustness of multi-class weed detection. To overcome these limitations, advanced deep learning architectures are essential for achieving reliable and efficient automated weeding under such complex conditions. Single-stage object detection methods based on convolutional neural networks (CNNs) have become the mainstream choice, with the YOLO series being the most representative \citep{terven2023comprehensive}.  By striking an optimal balance between detection accuracy and inference speed, YOLO models are highly capable of meeting the real-time detection demands of dynamic field environments \citep{ahmad2021performance}. In recent years, numerous studies \citep{sulzbach2025deep, gautam2025real, khan2025advancing} have applied models, such as YOLOv8 \citep{Jocher_Ultralytics_YOLO_2023}, to weed recognition tasks, achieving highly reliable performance. Building upon this foundation, the YOLO series has continuously evolved to further enhance feature extraction capabilities and inference efficiency. Successive developments include YOLOv9 \citep{wang2024yolov9}, the introduction of an end-to-end detection mechanism without Non-Maximum Suppression (NMS) in YOLOv10 \citep{wang2024yolov10}, the structurally optimized YOLOv11 \citep{jocher2024yolov11}, and the recent YOLOv12 \citep{tian2025yolov12}, which incorporates a region attention mechanism. With the continuous evolution of YOLO architectures, their applications in intelligent weed detection have become widespread. For instance, \cite{balasingham2024sparrow} deployed a YOLOv8 for real-time robotic spot-spraying. Also, \cite{saltik2024comparative} utilized standard YOLOv9 and YOLOv10 to establish highly efficient, low-latency detection baselines.

YOLO models are highly effective for object detection \citep{saltik2024comparative}, but their fundamental dependence on local receptive fields often leads to missed detections in complex agricultural environments with small targets \citep{zhang2022agripest} and heavy crop occlusion \citep{zhou2024efficient}. 
In recent years, the rapid development of vision Transformers has provided a new solution to overcome this limitation \citep{carion2020end}. As the first end-to-end Transformer-based object detection model capable of real-time inference, Real-Time Detection Transformer (RT-DETR) effectively alleviates the high computational cost of Transformers by introducing an efficient hybrid encoder \citep{zhao2024detrs, lv2024rt}. The model also eliminates the time-consuming NMS post-processing step and demonstrates excellent performance in multi-object real-time detection tasks under complex farmland environments \citep{allmendinger2025assessing}.   Building on these advantages, RT-DETR has significantly advanced the state-of-the-art in weed detection tasks. Building on these advantages, RT-DETR has significantly advanced the state-of-the-art in weed detection tasks \citep{gomez2025intelligent, garcia2025performance,saltik2024comparative}.

% 1 real environment deployment on edge device
Despite the significant progress achieved by deep learning–based object detection models in weed recognition and detection tasks in recent years, the training and evaluation of these models typically rely on high-performance computing environments \citep{srinivas2021bottleneck}. However, in practical agricultural applications, platforms such as spraying drones are constrained by limited payload capacity, power availability, and network connectivity, limiting the integration of high-performance computing hardware \citep{arsenoaia2026sensing}. Consequently, these small platforms rely on edge computing devices to perform real-time inference locally, leveraging embedded systems designed for low-power and resource-constrained environments \citep{adhikari2026comprehensive}.

% 2 trade-off accuracy and FPS 
In recent years, growing research attention has been directed toward validating deep learning models on edge devices in weed recognition and detection, with a particular focus on achieving real-time performance under limited computational resources \citep{upadhyay2024development, islam2025towards, yang2026weedcam, gauttam2026comprehensive}. 
For instance, \cite{rai2024agricultural} proposed the YOLO-Spot model, based on the YOLOv7-tiny architecture, which was optimized for weed detection in aerial images and videos. By converting the $\text{YOLO-Spot}\_\text{M}$ model to half-precision (FP16) for deployment on a resource-constrained platform (NVIDIA Jetson AGX Xavier), their approach achieved a 0.6\% improvement in detection accuracy and approximately a 5× increase in inference speed. Similarly, \cite{yang2026weedcam} developed a low-cost, near real-time edge-computing system based on the Raspberry Pi 5 for multi-species weed detection. In addition, \cite{islam2025towards} deployed lightweight models such as YOLOv8n and MobileNetV4-Seg on platforms like the Jetson Orin Nano, demonstrating the feasibility of real-time inference under constrained hardware settings. 
However, these studies are typically limited to evaluations on a single edge device and a narrow set of lightweight models, which restricts their ability to provide comprehensive insights into how different model architectures perform across diverse hardware platforms.

To bridge this gap, this study presents a deployment-oriented framework for systematically evaluating real-time weed detection models under edge constraints. Specifically, we conduct a comprehensive benchmarking study across multiple generations of convolutional models (YOLOv8-YOLOv12) and Transformer-based models (RT-DETRv1-v2), deployed on representative edge platforms including NVIDIA Jetson Orin Nano, Jetson AGX Xavier, and Jetson AGX Orin. Through this framework, we characterize the trade-offs between detection accuracy, inference latency, and model efficiency, providing practical guidelines for model selection in real-world agricultural applications.
A closely related study by \cite{ram2025edge} evaluated YOLOv8-YOLOv12 models for Palmer amaranth detection across multiple edge devices, demonstrating the feasibility of real-time deployment. In contrast, our work extends beyond a single model family by incorporating both convolutional and Transformer-based architectures and emphasizes cross-architecture comparison to provide more generalizable insights into model performance under diverse edge constraints.

\section{Materials and Methods}\label{sec:methods}
This section first describes the dataset used in this study, followed by an overview of the two model architectures evaluated, namely YOLO-based convolutional models and Transformer-based detection models. Finally, the experimental setup, including training procedures, evaluation metrics, and edge deployment configurations, is presented.

\subsection{Data acquisition}
The dataset used in this study was derived from the UAV imagery dataset reported in \citep{betitame2025practical}. The data were collected during the summer of 2022 at the Casselton Agronomy Seed Farm (CASF) in North Dakota, USA. The experimental field consisted of XtendFlex soybean, and aerial imagery was acquired during the early vegetative growth stages (V2–V3). This timing corresponds to the critical early-season weed control window, during which weed competition has the greatest impact on yield and management interventions are most effective.

\begin{figure}
    \centering
    \includegraphics[width=0.45\textwidth]{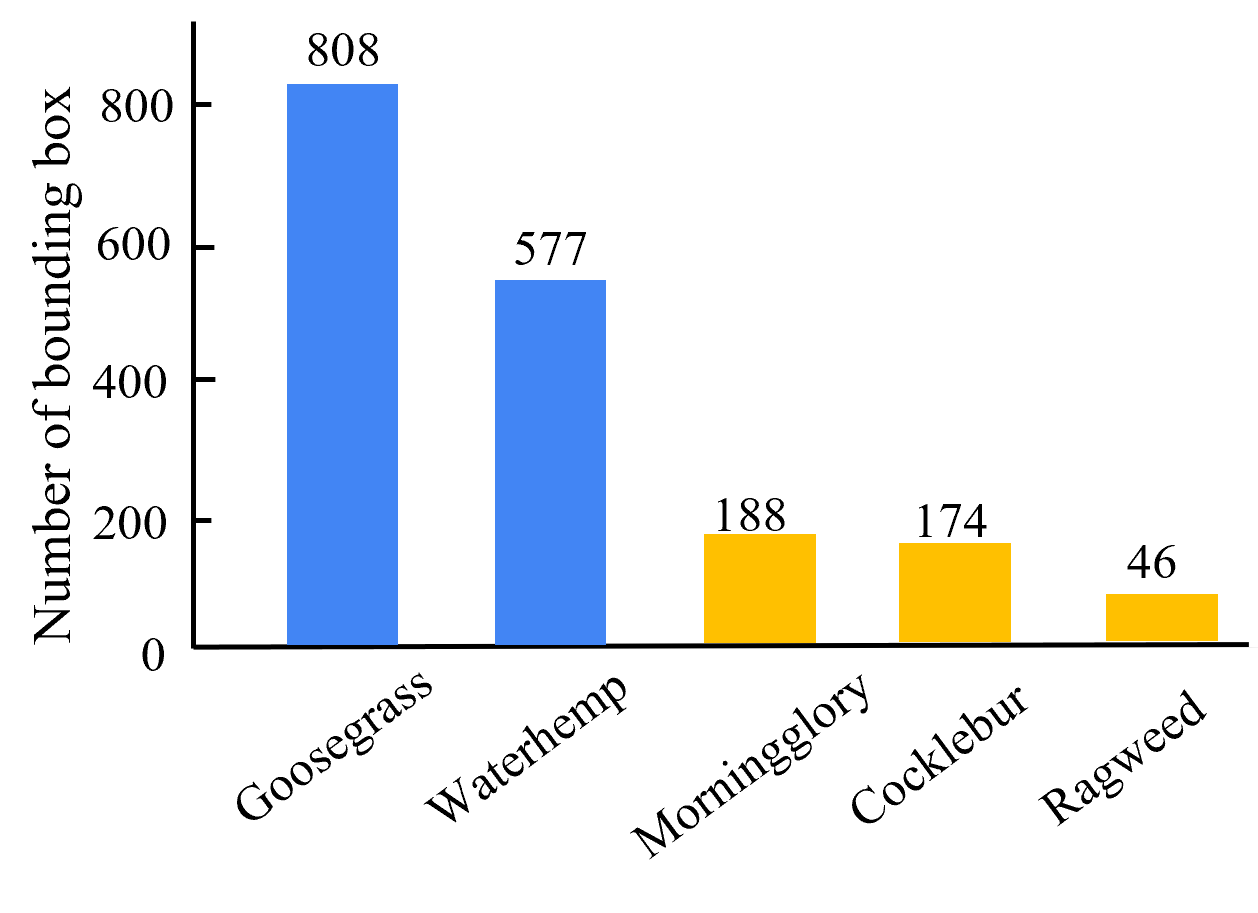} 
    \caption{Class distribution of weed species in the dataset.}
    \label{fig:weed_count}
\end{figure}

RGB images were captured using a DJI Phantom 4 Pro UAV with a resolution of 5472 × 3648 pixels. During the data collection process, the UAV operated at an altitude of approximately 10 m above ground level, with about 75\% front and side overlap between adjacent images. All images were captured using a mechanical shutter, which minimizes motion blur and preserves clear morphological details of crops and weeds, making the dataset suitable for weed detection tasks.
Fig.~\ref{fig:weed_count} illustrates the class distribution of the dataset. The dataset is notably imbalanced, with Goosegrass (808 samples) and Waterhemp (577 samples) dominating the dataset, while Morningglory (188 samples), Cocklebur (174 samples), and Ragweed (46 samples) are underrepresented. The selected weed species represent a mixture of grasses and broadleaf weeds commonly encountered in soybean systems, many of which differ in competitiveness and herbicide response, making accurate classification particularly important for guiding appropriate management strategies.

\subsection{YOLO architecture models}
The YOLO framework formulates object detection as a single-stage regression problem, enabling the model to directly predict object locations and categories from an input image. The architecture generally consists of three major components: a backbone network for hierarchical feature extraction, a neck module for aggregating features across multiple spatial scales, and a detection head that outputs bounding box coordinates along with class probabilities \citep{vijayakumar2024yolo}. During training, the model is optimized using a composite loss function that jointly accounts for localization accuracy, classification performance, and objectness confidence. During inference, the network processes the entire image in a single forward pass and applies NMS to remove redundant overlapping predictions, producing the final detection results efficiently.
This single-stage design makes YOLO particularly suitable for real-time applications, especially in resource-constrained edge environments \citep{adhikari2026comprehensive}. 

The following subsections present the YOLO variants (v8–v12). Earlier versions are excluded to ensure a fair and consistent comparison, as YOLOv8 and subsequent models share more unified design principles and optimization strategies tailored for modern real-time and edge deployment scenarios.

\subsubsection{YOLOv8}
YOLOv8 represents a significant advancement in object detection architecture and addresses several limitations of previous YOLO variants. It adopts an anchor-free detection framework in place of the traditional anchor-based design, reducing reliance on manual design choices and simplifying the detection pipeline \citep{Jocher_Ultralytics_YOLO_2023}. The backbone and feature fusion network are improved with a more efficient design that enhances gradient propagation and strengthens multi-scale feature representation, making it effective for detecting objects with significant scale variation. YOLOv8 also introduces improved training strategies for dynamic sample assignment and more accurate bounding box regression, which collectively enhance localization performance.

\subsubsection{YOLOv9} 
YOLOv9 addresses the information bottleneck problem in deep neural networks by introducing improved feature aggregation and gradient propagation mechanisms \citep{wang2024yolov9}. It adopts a more efficient network design that enhances multi-scale feature representation while maintaining computational efficiency, enabling the model to better preserve fine-grained spatial information. A training strategy based on an auxiliary reversible branch is employed to facilitate stable gradient flow, improving optimization quality and localization accuracy without increasing inference-time computational cost.

\subsubsection{YOLOv10}
YOLOv10 introduces an NMS-free detection framework to streamline the inference pipeline and improve real-time performance \citep{wang2024yolov10}. It adopts an end-to-end training strategy that ensures stable convergence and consistent prediction quality. The network is further optimized with efficient architectural designs that reduce computational overhead and preserve important spatial information for feature representation. These improvements lead to lower inference latency and make the model well-suited for real-time deployment on resource-constrained devices.

\subsubsection{YOLOv11}
YOLOv11 focuses on structural optimizations to achieve superior parameter efficiency and faster inference speeds \citep{jocher2024yolov11}. By refining its backbone and feature fusion layers, the model improves gradient flow and reduces computational redundancy. A significant advancement in YOLOv11 is the official integration of a self-attention mechanism, which enhances the network's ability to capture long-range spatial dependencies. This capability is particularly beneficial for distinguishing small weed seedlings from complex backgrounds, such as soil and crop residues, making YOLOv11 well-suited for edge-based agricultural applications.

\subsubsection{YOLOv12}
Building upon these improvements, YOLOv12 introduces an attention-centric architecture designed to maximize feature representation while maintaining real-time performance \citep{tian2025yolov12}. Unlike previous versions that used attention as a supplementary component, YOLOv12 centers its design around an efficient attention mechanism that captures long-range spatial dependencies more holistically. This architectural shift enables the model to aggregate multi-scale features more effectively, leading to more stable training and improved localization accuracy. For UAV-based weed mapping, these enhancements ensure consistent detection performance across diverse field conditions and varying weed growth stages.

\subsection{Transformer architecture models}
While most YOLO models exhibit exceptional real-time capabilities, their reliance on NMS for post-processing presents a fundamental limitation in densely populated scenes. NMS often incorrectly suppresses adjacent bounding boxes, which is a critical drawback when detecting densely clustered weeds in UAV imagery. To address this, RT-DETR models offer a fully end-to-end alternative \citep{zhao2024detrs}. In RT-DETR, multi-scale features are first extracted using a backbone network and then processed by a Transformer-based encoder–decoder architecture. A bipartite matching strategy is employed to directly generate final predictions without the need for post-processing steps such as NMS. By eliminating the NMS stage, RT-DETR reduces the risk of removing overlapping targets and simplifies the inference pipeline. This design makes it particularly suitable for dense agricultural scenarios, where accurate detection of closely spaced objects is critical \citep{xue2025real}.

\subsubsection{RT-DETRv1}
RT-DETRv1 represents the first model in this series to demonstrate that 
Transformer-based object detection can be both accurate and practical under real-time constraints \citep{zhao2024detrs}. To address the high computational burden typically associated with Transformer detectors, it introduces an efficient hybrid encoder that separates intra-scale feature learning from cross-scale feature interaction. This design reduces computational complexity while retaining important spatial details. Furthermore, an Intersection over Union (IoU)-aware query selection strategy is adopted to initialize the decoder with higher-quality object queries, allowing the model to focus on more relevant candidate regions at an early stage. These design choices lead to stronger localization performance, especially for small or less salient objects, while maintaining real-time inference efficiency.

\subsubsection{RT-DETRv2}
RT-DETRv2 builds on this framework with several refinements to improve feature representation and architectural flexibility \citep{lv2024rt}. Notably, it introduces deformable attention in the decoder, enabling adaptive sampling across spatial locations and scales. This allows the model to better capture fine-grained details, which is especially advantageous for UAV-based weed detection where targets vary in size, shape, and distribution. Additionally, enhanced training strategies improve convergence stability and robustness without compromising real-time inference performance.

Table~\ref{tab:detector_models} summarizes the representative real‑time object detection models evaluated in this paper, including the YOLOv8-YOLOv12 and RT‑DETRv1\&v2  series, along with their corresponding core references and publicly available open‑source GitHub repositories. Note that each model family contains multiple variants (e.g., $n$, $s$, $m$, $l$, $x$), which represent different model scales in terms of network depth, width, and computational complexity, thereby offering trade-offs between accuracy and efficiency. 

% \begin{table}[ht]
% \centering
% \caption{Summary of recent real-time object detection models evaluated in this paper.}
% \label{tab:detector_models}
% \resizebox{\columnwidth}{!}{
% \begin{tabular}{clll} 
% \toprule
% \textbf{Index} & \textbf{Models} & \textbf{GitHub links} & \textbf{References} \\
% \midrule
% 1 & YOLOv8  & \href{https://github.com/ultralytics/ultralytics}{\makecell[l]{github.com/ultralytics/\\ultralytics}} & \cite{Jocher_Ultralytics_YOLO_2023}  \\
% 2 & YOLOv9  & \href{https://github.com/WongKinYiu/yolov9}{\makecell[l]{github.com/WongKinYiu/\\yolov9}}       & \cite{wang2024yolov9}    \\
% 3 & YOLOv10 & \href{https://github.com/THU-MIG/yolov10}{\makecell[l]{github.com/THU-MIG/\\yolov10}}         & \cite{wang2024yolov10}   \\
% 4 & YOLOv11 & \href{https://github.com/ultralytics/ultralytics}{\makecell[l]{github.com/ultralytics/\\ultralytics}} & \cite{jocher2024yolov11} \\
% 5 & YOLOv12 & \href{https://github.com/sunsmarterjie/yolov12}{\makecell[l]{github.com/sunsmarterjie/\\yolov12}}   & \cite{tian2025yolov12}   \\
% 6 & RT-DETR-v1 & \href{https://github.com/lyuwenyu/RT-DETR}{\makecell[l]{github.com/lyuwenyu/\\RT-DETR}}     & \cite{zhao2024detrs}     \\
% 7 & RT-DETR-v2 & \href{https://github.com/lyuwenyu/RT-DETR}{\makecell[l]{github.com/lyuwenyu/\\RT-DETR}}     & \cite{lv2024rt}          \\
% \bottomrule
% \end{tabular}
% }
% \end{table}
\begin{table*}[htbp]
\centering
\caption{Summary of recent real-time object detection models evaluated in this paper, including their key architectural features.}
\label{tab:detector_models}
\setlength{\tabcolsep}{12pt} 
\begin{tabular}{c l l l} 
\toprule
\textbf{Index} & \textbf{Models} & \textbf{Architectural Features} & \textbf{GitHub Repository} \\
\midrule
1 & YOLOv8 \citep{Jocher_Ultralytics_YOLO_2023}  & Anchor-free, C2f module & \href{https://github.com/ultralytics/ultralytics}{\ttfamily ultralytics/ultralytics} \\
\addlinespace
2 & YOLOv9 \citep{wang2024yolov9} & PGI \& GELAN blocks & \href{https://github.com/WongKinYiu/yolov9}{\ttfamily WongKinYiu/yolov9} \\
\addlinespace
3 & YOLOv10 \citep{wang2024yolov10} & NMS-free dual assignments & \href{https://github.com/THU-MIG/yolov10}{\ttfamily THU-MIG/yolov10} \\
\addlinespace
4 & YOLOv11 \citep{jocher2024yolov11} & C3k2 \& C2PSA modules & \href{https://github.com/ultralytics/ultralytics}{\ttfamily ultralytics/ultralytics} \\
\addlinespace
5 & YOLOv12 \citep{tian2025yolov12} & Attention-centric design & \href{https://github.com/sunsmarterjie/yolov12}{\ttfamily sunsmarterjie/yolov12} \\
\addlinespace
6 & RT-DETRv1 \citep{zhao2024detrs} & IoU-guided Query & \href{https://github.com/lyuwenyu/RT-DETR}{\ttfamily lyuwenyu/RT-DETR} \\
\addlinespace
7 & RT-DETRv2 \citep{lv2024rt} &  Adaptive Query Refinement & \href{https://github.com/lyuwenyu/RT-DETR}{\ttfamily lyuwenyu/RT-DETR} \\
\bottomrule
\end{tabular}
\end{table*}

\subsection{Model training \& evaluation pipeline} \label{sec:sub:training}
\begin{figure*}[htbp]
    \centering
    \includegraphics[width=0.95\textwidth]{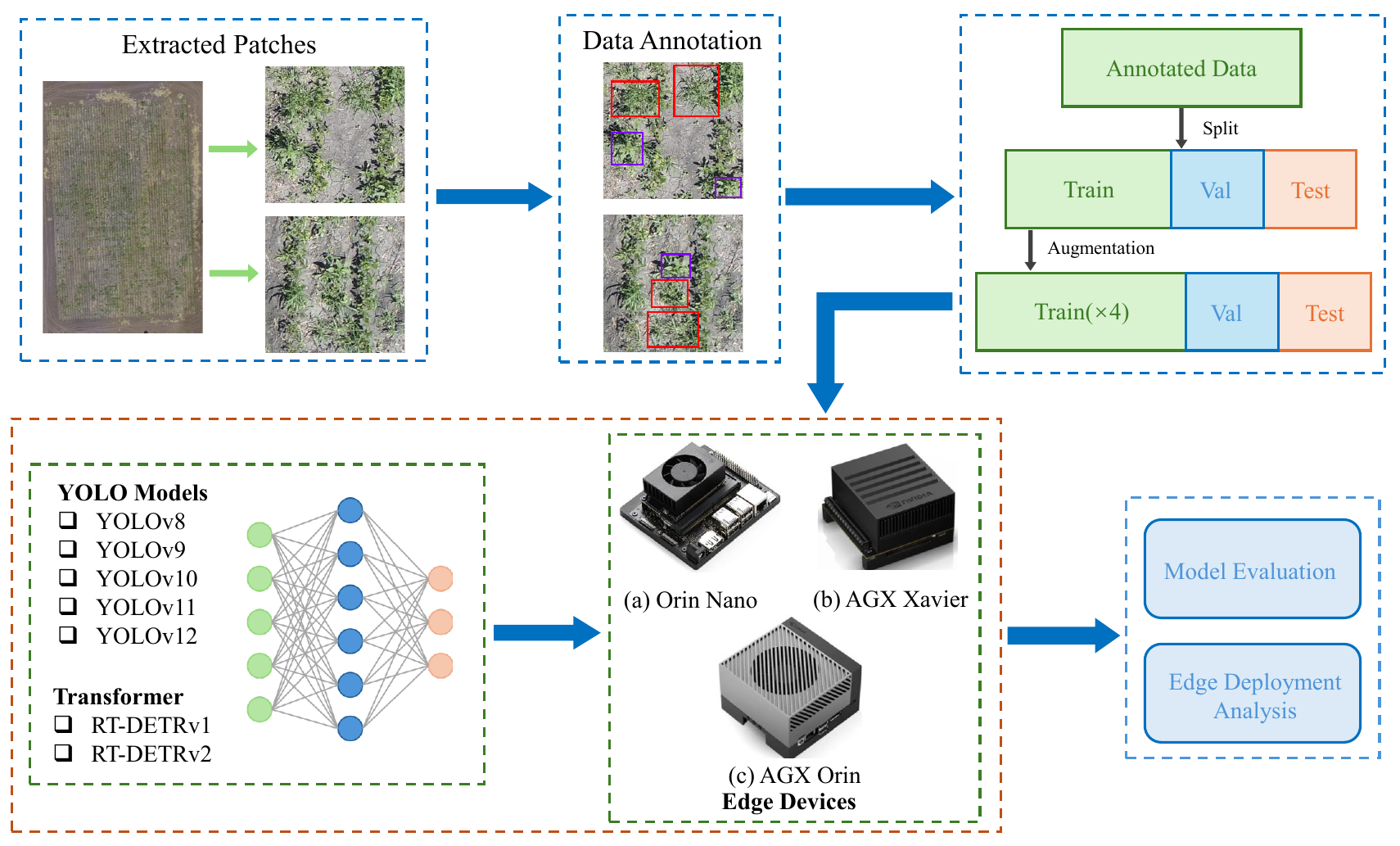} 
    \caption{Overview of the experimental pipeline for evaluating object detection models on edge computing platforms.}
    \label{fig:pipeline}
\end{figure*}

Fig.~\ref{fig:pipeline} illustrates the overall pipeline for the proposed weed detection system.  Standard UAV data acquisition requires high frontal and sidelap, which introduces substantial redundancy and repeated weed instances in the raw image set. To eliminate this redundancy and enable accurate spatial analysis, the collected UAV images were first processed in Pix4D to generate a georeferenced orthomosaic of the soybean field. Ground control points (GCPs) were incorporated during this process to ensure precise spatial alignment with real‑world coordinates, thereby supporting consistent and repeatable weed mapping \citep{betitame2025practical}. The resulting orthomosaic, featuring a spatial resolution of 0.25~cm/pixel, was then partitioned into $640 \times 640$ pixel patches with a 10\% overlap. This preprocessing strategy effectively resolves the excessive redundancy of the initial capture while preserving high‑resolution spatial information critical for discriminating weeds from crops.

The annotated dataset was randomly split into training (80\%), validation (10\%), and testing (10\%) subsets. To improve model robustness and generalization, four data augmentation techniques, e.g., flipping, shearing, cropping, and mosaic augmentation, were applied exclusively to the training set prior to model training and evaluation \citep{bochkovskiy2020yolov4}. Following model training, the trained object detection networks were deployed on multiple NVIDIA‑based edge computing platforms, including Jetson Orin Nano, Jetson AGX Xavier, and Jetson AGX Orin, as illustrated in Fig.~\ref{fig:pipeline}. Each model was optimized for edge inference and evaluated under real‑world deployment constraints.

% Edge device introduction
\subsection{Inference on edge devices}
In autonomous agricultural scenarios, UAVs must balance real-time processing with strict onboard power and computational constraints. Consequently, we prioritized small and medium-sized model variants to ensure both high detection efficiency and prolonged flight endurance on edge-deployed devices. Three representative NVIDIA Jetson platforms were selected for evaluation:  Jetson Orin Nano, Jetson AGX Xavier, and Jetson AGX Orin. All platforms leverage NVIDIA TensorRT for inference acceleration, enabling optimization techniques such as layer fusion and precision quantization. These optimizations significantly improve computational throughput and real-time inference performance while operating within constrained power envelopes. The detailed hardware configurations of the evaluated platforms are summarized in Table~\ref{tab:hardware_specs}.

All YOLO and RT-DETR models were first trained and validated on a High-Performance Computing (HPC) server to obtain their best-performing weights. The optimized model checkpoints were then deployed to the edge platforms for inference evaluation. To ensure compatibility with embedded hardware and reduce inference latency, the trained models were converted to the ONNX format and executed with FP16 precision on the Jetson devices.

\subsubsection{Jetson Orin Nano}
The Jetson Orin Nano provides a lightweight and cost-effective edge computing solution with a compact form factor and low power consumption, making it well suited for small UAVs and ground robotic platforms. However, its limited computational capacity restricts the deployment of larger and more computationally intensive models. Experiments on this platform focus on analyzing the trade-off between detection accuracy and inference efficiency under realistic resource constraints.

\subsubsection{Jetson AGX Xavier}
The Jetson AGX Xavier is included as a previous-generation flagship platform to facilitate comparative analysis across different edge computing architectures. Although it offers 32~GB of system memory, its Volta-based GPU architecture exhibits lower inference efficiency compared to newer Ampere-based platforms. As a result, the AGX Xavier serves as a baseline reference for assessing the performance and deployment feasibility of state-of-the-art detection models on legacy edge hardware.

\subsubsection{Jetson AGX Orin}
The Jetson AGX Orin represents the upper bound of current edge computing capabilities, featuring an NVIDIA Ampere architecture GPU and delivering up to 275 TOPS of AI performance. In this study, the AGX Orin serves as a high-performance benchmark platform to evaluate the maximum achievable inference accuracy and throughput of complex detection models. Its performance establishes a reference level for near-optimal real-time deployment in agricultural UAV applications.

\begin{table*}[t]
    \centering
    \caption{Hardware specifications of the development server and evaluated NVIDIA Jetson edge computing platforms.}
    \label{tab:hardware_specs}
    \renewcommand{\arraystretch}{1.15}
    \begin{tabularx}{\textwidth}{lXXcc}
    \toprule
    \textbf{Platform} & \textbf{CPU} & \textbf{GPU} & \textbf{RAM} & \textbf{Power} \\
    \midrule
    \textbf{Training Server} &
    24-core AMD Ryzen Threadripper PRO 7965WX (64‑bit) &
    NVIDIA GeForce RTX 5090 &
    128\,GB &
    600–1500\,W \\
    \addlinespace
    
    \textbf{Jetson Orin Nano} &
    6-core Arm Cortex‑A78AE (v8.2, 64‑bit) &
    1024-core NVIDIA Ampere GPU with 32 Tensor Cores &
    8\,GB &
    7–15\,W \\
    \addlinespace
    
    \textbf{Jetson AGX Xavier} &
    8-core NVIDIA Carmel Arm (v8.2, 64‑bit) &
    512-core NVIDIA Volta GPU with Tensor Cores &
    32\,GB &
    10–30\,W \\
    \addlinespace
    
    \textbf{Jetson AGX Orin} &
    12-core Arm Cortex‑A78AE (v8.2, 64‑bit) &
    2048-core NVIDIA Ampere GPU with 64 Tensor Cores &
    64\,GB &
    15–60\,W \\
    \bottomrule
    \end{tabularx}
\end{table*}

\subsection{Evaluation metrics}
To quantitatively assess the performance of the selected YOLO and RT-DETR models, a standard set of evaluation metrics was employed to capture both detection accuracy and localization quality. The evaluated metrics include Precision, Recall, F1-score, and mean Average Precision (mAP). 

\subsubsection{Precision and Recall}
In object detection, performance evaluation is fundamentally based on the IoU between predicted bounding boxes and ground-truth annotations. Given a predefined IoU threshold, detection outcomes are categorized as True Positives (TP), False Positives (FP), or False Negatives (FN), which form the basis for all subsequent metric calculations \citep{lin2014microsoft}.
Precision measures the proportion of correctly detected positive instances among all predicted positives, reflecting the detector’s ability to avoid false alarms. Recall quantifies the proportion of actual positive instances that are successfully detected, indicating the detector’s sensitivity. These metrics are defined as:
\begin{equation}
\text{Precision} = \frac{TP}{TP + FP}
\end{equation}
\begin{equation}
\text{Recall} = \frac{TP}{TP + FN}
\end{equation}

\subsubsection{F1-score}
Since Precision and Recall often exhibit a trade-off, the F1-score is used as their harmonic mean to provide a balanced measure of detection performance for each class $i$. To assess the overall performance across all weed categories, the Macro-F1 score is computed as the arithmetic mean of the class-wise F1-scores:
\begin{equation}
F1_{i} = 2 \times \frac{\text{Precision}_{i} \times \text{Recall}_{i}}{\text{Precision}_{i} + \text{Recall}_{i}}
\end{equation}

\begin{equation}
\text{Macro-F1} = \frac{1}{N} \sum_{i=1}^{N} F1_{i}
\end{equation}
where $N$ denotes the total number of evaluated classes. In the following sections, all reported F1-scores correspond to the Macro-F1 score.

\subsubsection{Mean Average Precision}
Mean Average Precision (mAP) serves as the primary metric for evaluating object detection performance \citep{everingham2010pascal}. It is calculated as the mean of the Average Precision (AP) values across all classes, where AP corresponds to the area under the Precision–Recall curve:
\begin{equation}
mAP = \frac{1}{C} \sum_{i=1}^{C} AP_i
\end{equation}

In this study, two commonly used variants are reported: $\text{mAP}_{50}$, computed at a single IoU threshold of 0.5, and $\text{mAP}_{50:95}$, which averages AP over IoU thresholds ranging from 0.5 to 0.95 with a step size of 0.05. The $\text{mAP}_{50:95}$ metric provides a more rigorous evaluation of both detection and localization accuracy.

\subsubsection{Model complexity and inference time}
Model complexity is a critical factor for real-world deployment, particularly on resource-constrained edge devices. The number of model parameters reflects memory requirements, while Giga Floating-Point Operations (GFLOPs) indicate the theoretical computational cost of a model. Larger models with higher GFLOPs generally impose greater computational loads, leading to increased power consumption, elevated chip temperatures, and longer inference latency. Although parameter count and GFLOPs provide useful indicators of model size and complexity, they do not fully capture practical runtime performance on embedded platforms \citep{ma2018shufflenet}. Therefore, end-to-end inference time is adopted as the primary efficiency metric in this study. This metric accounts for the complete processing pipeline, including input preprocessing, network inference, and post-processing, and thus provides a more realistic assessment of deployment feasibility for real-time agricultural applications.

\begin{figure*}[htbp]
    \centering
    \includegraphics[width=0.95\linewidth]{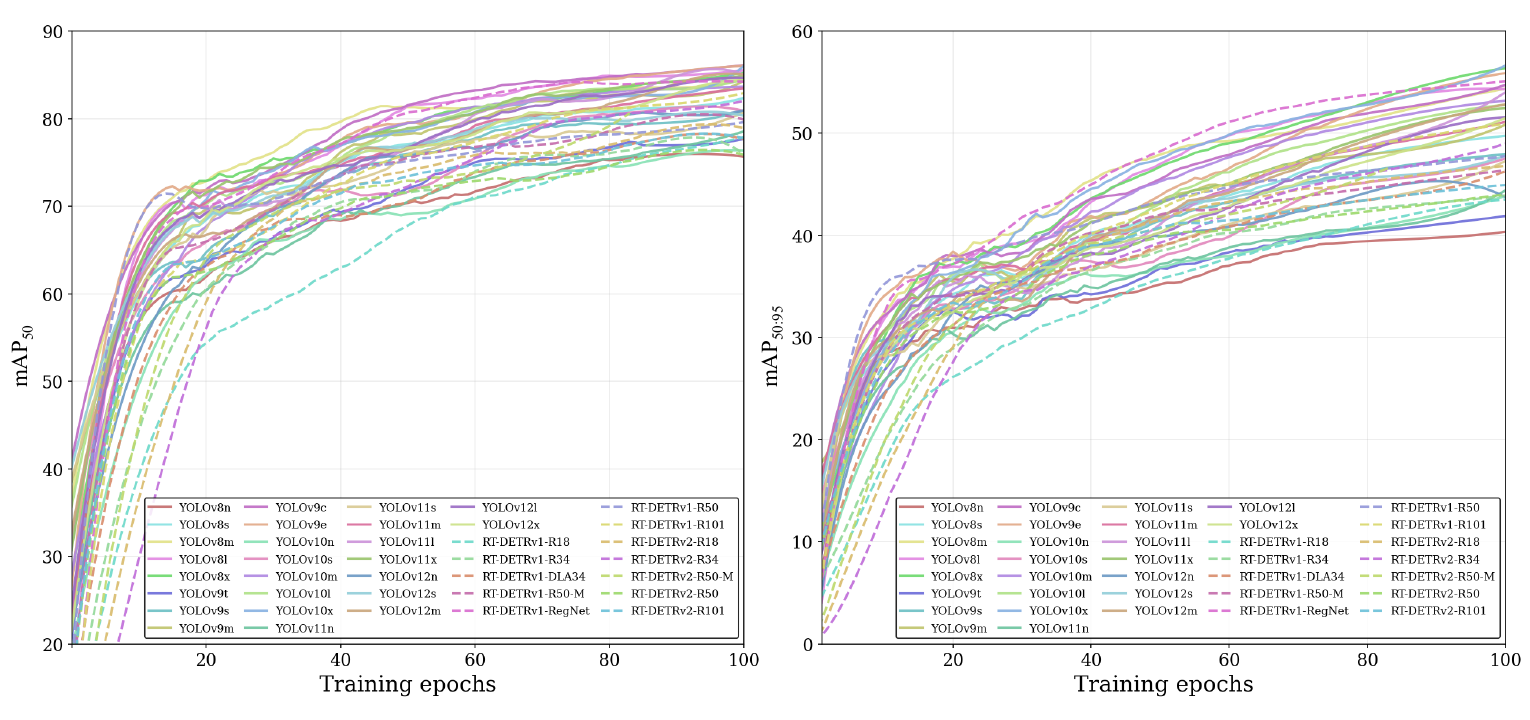} 
    \caption{Training accuracy curves of YOLO and RT-DETR models with data augmentation.}
    \label{fig:training}
\end{figure*}

\section{Experiments and Results}\label{sec:results}

\subsection{Performance of overall models}
Fig.~\ref{fig:training} illustrates the convergence behavior of $\text{mAP}_{50}$ and $\text{mAP}_{50:95}$ over 100 training epochs for the YOLO series (solid lines) and RT-DETR variants (dashed lines). All evaluated models demonstrate stable and consistent convergence trends. Specifically, rapid performance improvements are observed during the early training stages, followed by a gradual saturation phase. In the final 10 epochs, the performance of all models stabilizes, with $\text{mAP}_{50}$ values generally ranging from approximately 75\% to 85\%.

Table~\ref{combined_models} presents a comprehensive comparison of all 37 evaluated object detection models. For clarity, the best-performing model within each series is highlighted in bold based on the $\text{mAP}_{50}$ metric. Overall, the RT-DETR family achieves a higher upper bound in detection accuracy compared to the YOLO counterparts. In particular, RT-DETR-RegNet delivers the best overall performance, achieving an $\text{mAP}_{50}$ of 86.93\% and an $\text{mAP}_{50:95}$ of 53.91\%, along with a strong F1-score of 85.32\%.
In contrast, within the single-stage YOLO family, YOLOv9e achieves the highest performance, with an $\text{mAP}_{50}$ of 84.90\% and an $\text{mAP}_{50:95}$ of 51.15\%. While YOLO models demonstrate competitive accuracy and generally lower computational complexity, RT-DETR models exhibit superior detection capability, particularly in achieving higher precision and recall simultaneously, as reflected by their stronger F1-scores.

\begin{table*}[htbp]
  \centering
  \caption{Performance comparison of YOLO and RT-DETR object detection models.}
  \label{combined_models}
  \renewcommand{\arraystretch}{1.1} 
  \resizebox{0.9\textwidth}{!}{%
  \begin{tabular}{c l l c c c c c c c}
    \toprule
    Index & \multicolumn{2}{c}{Models} & Precision & Recall & F1 & mAP$_{50}$ & mAP$_{50:95}$ & Params (M) & GFLOPs \\
    \midrule
    % YOLOv8 Group
    1 & \multirow{5}{*}{YOLOv8} & YOLOv8n & 81.56 & 65.52 & 72.25 & 72.22 & 38.15 & 3.16 & 8.86 \\
    2 & & YOLOv8s & 78.18 & 77.59 & 76.93 & 75.29 & 43.65 & 11.17 & 28.82 \\
    3 & & YOLOv8m & 84.62 & 72.62 & 78.06 & 76.46 & 45.84 & 25.90 & 79.32 \\
    4 & & \textbf{YOLOv8l} & \textbf{80.19} & \textbf{80.92} & \textbf{80.12} & \textbf{82.42} & \textbf{51.80} & 43.69 & 165.74 \\
    5 & & YOLOv8x & 79.63 & 81.28 & 80.18 & 80.16 & 50.01 & 68.23 & 258.55 \\ 
    \addlinespace[0.2ex] \midrule \addlinespace[0.2ex]

    % YOLOv9 Group
    6 & \multirow{5}{*}{YOLOv9} & YOLOv9t & 74.96 & 71.64 & 72.77 & 69.62 & 36.38 & 2.13 & 8.48 \\
    7 & & YOLOv9s & 80.77 & 76.82 & 78.37 & 77.95 & 46.47 & 7.32 & 27.56 \\
    8 & & YOLOv9m & 82.77 & 74.48 & 77.75 & 78.77 & 47.89 & 20.22 & 77.87 \\
    9 & & YOLOv9c & 83.27 & 80.68 & 81.80 & 83.19 & 52.25 & 25.59 & 104.02 \\
    10 & & \textbf{YOLOv9e} & \textbf{84.56} & \textbf{81.88} & \textbf{82.77} & \textbf{84.90} & \textbf{51.15} & 58.21 & 193.02 \\
    \addlinespace[0.2ex] \midrule \addlinespace[0.2ex]

    % YOLOv10 Group
    11 & \multirow{5}{*}{YOLOv10} & YOLOv10n & 75.72 & 73.76 & 74.33 & 75.67 & 41.49 & 2.78 & 8.74 \\
    12 & & YOLOv10s & 80.80 & 68.92 & 73.55 & 75.91 & 41.68 & 8.13 & 25.11 \\
    13 & & YOLOv10m & 78.22 & 79.62 & 77.95 & 79.54 & 47.47 & 16.58 & 64.48 \\
    14 & & YOLOv10l & 80.40 & 77.49 & 78.16 & 81.42 & 47.78 & 25.89 & 127.87 \\
    15 & & \textbf{YOLOv10x} & \textbf{83.24} & \textbf{78.76} & \textbf{80.01} & \textbf{82.10} & \textbf{50.17} & 31.81 & 171.85 \\
    \addlinespace[0.2ex] \midrule \addlinespace[0.2ex]

    % YOLOv11 Group
    16 & \multirow{5}{*}{YOLOv11} & YOLOv11n & 73.28 & 74.10 & 72.85 & 70.47 & 38.20 & 2.62 & 6.61 \\
    17 & & YOLOv11s & 86.07 & 73.93 & 79.27 & 81.01 & 43.94 & 9.46 & 21.72 \\
    18 & & YOLOv11m & 80.09 & 81.49 & 80.70 & 80.86 & 46.94 & 20.11 & 68.53 \\
    19 & & \textbf{YOLOv11l} & \textbf{84.21} & \textbf{78.94} & \textbf{80.93} & \textbf{83.26} & \textbf{49.84} & 25.37 & 87.61 \\
    20 & & YOLOv11x & 86.83 & 72.69 & 78.72 & 79.18 & 47.01 & 56.97 & 195.96 \\
    \addlinespace[0.2ex] \midrule \addlinespace[0.2ex]

    % YOLOv12 Group
    21 & \multirow{5}{*}{YOLOv12} & YOLOv12n & 72.14 & 78.98 & 74.71 & 75.08 & 39.03 & 2.60 & 6.65 \\
    22 & & YOLOv12s & 79.09 & 83.40 & 81.00 & 82.47 & 47.86 & 9.29 & 21.69 \\
    23 & & YOLOv12m & 86.60 & 76.71 & 80.61 & 82.66 & 48.95 & 20.20 & 68.08 \\
    24 & & YOLOv12l & 79.45 & 76.24 & 77.33 & 80.23 & 45.22 & 26.45 & 89.75 \\
    25 & & \textbf{YOLOv12x} & \textbf{78.51} & \textbf{81.55} & \textbf{79.67} & \textbf{83.21} & \textbf{48.33} & 59.22 & 200.33 \\
    \addlinespace[0.2ex] \midrule \addlinespace[0.2ex]

    % RT-DETR v1 Group
    26 & \multirow{7}{*}{RT-DETRv1} & RT-DETRv1-R18 & 76.41 & 66.89 & 70.61 & 71.33 & 36.31 & 20.09 & 60.43 \\
    27 & & RT-DETRv1-R34 & 69.82 & 76.22 & 72.49 & 74.65 & 41.61 & 31.33 & 92.47 \\
    28 & & RT-DETRv1-DLA34 & 71.77 & 77.16 & 74.06 & 76.42 & 43.84 & 33.80 & 113.25 \\
    29 & & RT-DETRv1-R50-M & 75.54 & 71.40 & 73.03 & 76.43 & 42.48 & 36.44 & 99.32 \\
    30 & & \textbf{RT-DETRv1-RegNet} & \textbf{85.82} & \textbf{85.16} & \textbf{85.32} & \textbf{86.93} & \textbf{53.91} & 38.36 & 129.17 \\
    31 & & RT-DETRv1-R50 & 83.90 & 75.44 & 79.23 & 81.14 & 47.02 & 42.74 & 135.82 \\
    32 & & RT-DETRv1-R101 & 80.31 & 80.58 & 80.03 & 83.08 & 46.06 & 76.40 & 258.46 \\
    \addlinespace[0.2ex] \midrule \addlinespace[0.2ex]

    % RT-DETR v2 Group
    33 & \multirow{5}{*}{RT-DETRv2} & RT-DETRv2-R18 & 83.19 & 76.65 & 79.08 & 80.82 & 46.36 & 20.09 & 60.43 \\
    34 & & \textbf{RT-DETRv2-R34} & \textbf{80.34} & \textbf{84.14} & \textbf{81.99} & \textbf{85.19} & \textbf{48.60} & 31.33 & 92.47 \\
    35 & & RT-DETRv2-R50-M & 75.73 & 77.75 & 76.06 & 79.06 & 44.93 & 36.44 & 99.32 \\
    36 & & RT-DETRv2-R50 & 74.16 & 75.51 & 73.17 & 74.68 & 39.42 & 42.74 & 135.82 \\
    37 & & RT-DETRv2-R101 & 79.11 & 79.19 & 78.96 & 80.32 & 45.21 & 76.40 & 258.46 \\
    \bottomrule
  \end{tabular}%
  }
\end{table*}

\begin{figure*}[htbp]
    \centering
    \includegraphics[width=0.99\textwidth]{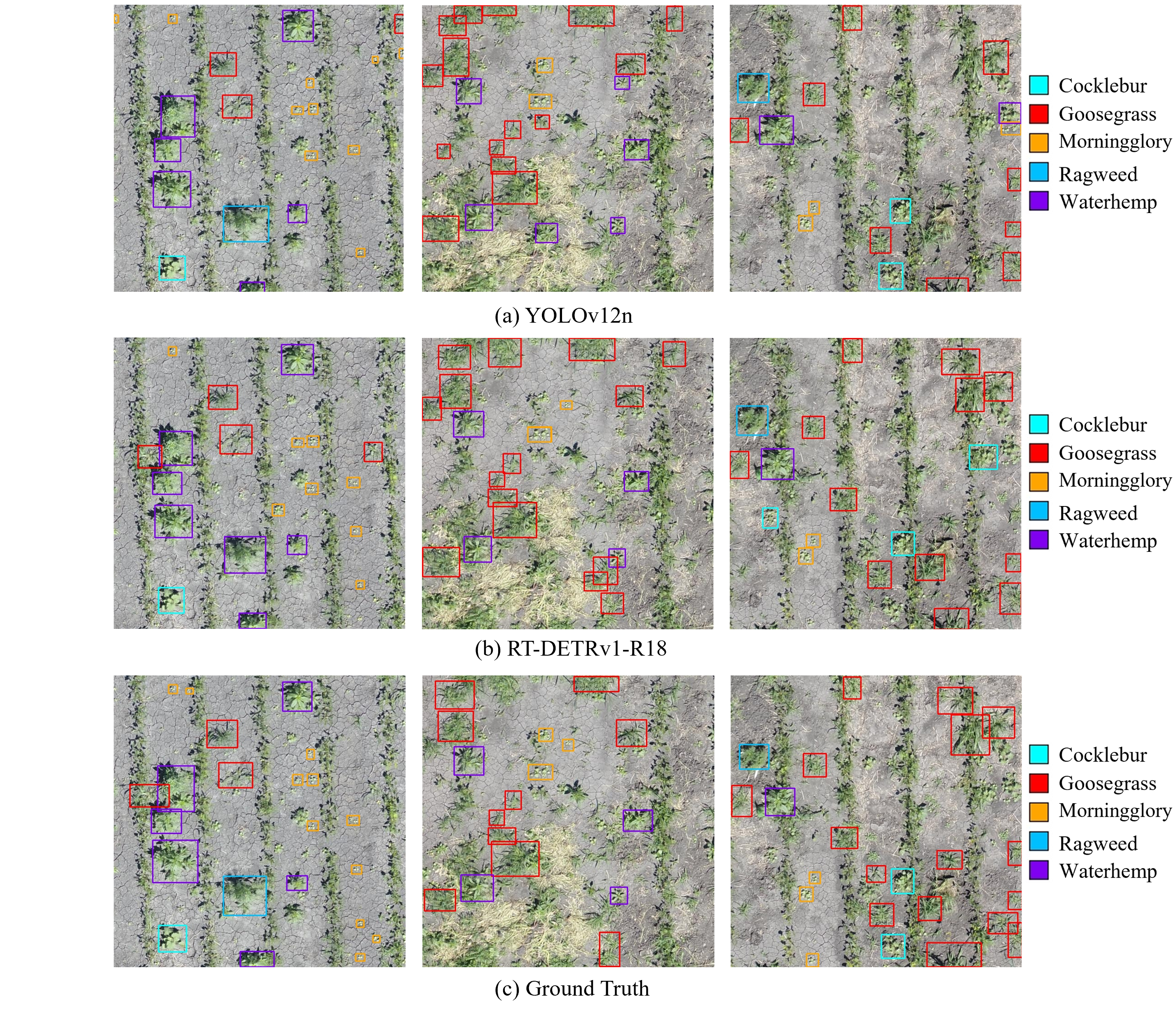} 
        \vspace{-10pt}
    \caption{Visual comparison of multi-class weed detection results using YOLOv12n and RT-DETRv1-R18.}
    \label{visual_result}
\end{figure*}

\subsection{Performance of individual weed categories}
We further evaluate detection performance at the per-class level. Considering the constraints of low inference latency and energy efficiency for UAV-based deployment, two lightweight models, i.e., YOLOv12n and RT-DETRv1-R18, are selected for detailed analysis. The quantitative results are summarized in Table~\ref{tab:weed_comparison}, while qualitative examples are shown in Fig.~\ref{visual_result}.
Overall, detection performance is highly influenced by object scale, spatial density, and inter-class morphological similarity. As a dominant class, Goosegrass enables both models to achieve stable and reliable performance, with $\text{mAP}_{50}$ values exceeding 80\% in both cases. In contrast, although Waterhemp is relatively abundant in the dataset, its detection accuracy remains limited. This can be attributed to distorted visual features under UAV imaging perspectives and motion-induced blurriness, both of which degrade image quality and hinder feature extraction.

% explain for the bad performance
Morningglory represents the most challenging class, exhibiting the lowest $\text{mAP}_{50}$ across both models. This is primarily due to its extremely small object size, often occupying only a few tens of pixels per instance, resulting in insufficient spatial information for robust detection. Notably, RT-DETRv1 demonstrates a clear advantage on this class, outperforming YOLOv12n by approximately 9\% in $\text{mAP}_{50}$, highlighting the benefit of its global attention mechanism in handling small and sparse targets.
In contrast, Ragweed presents an interesting case. Despite being a minority class, it achieves the highest detection accuracy, reaching 95.05\% $\text{mAP}_{50}$ with YOLOv12n. This strong performance is likely due to its distinctive visual characteristics, which make it easier to differentiate from both background and other weed species, even with limited training samples.

\begin{figure*}[t]  
    \centering
    \includegraphics[width=0.9\textwidth]{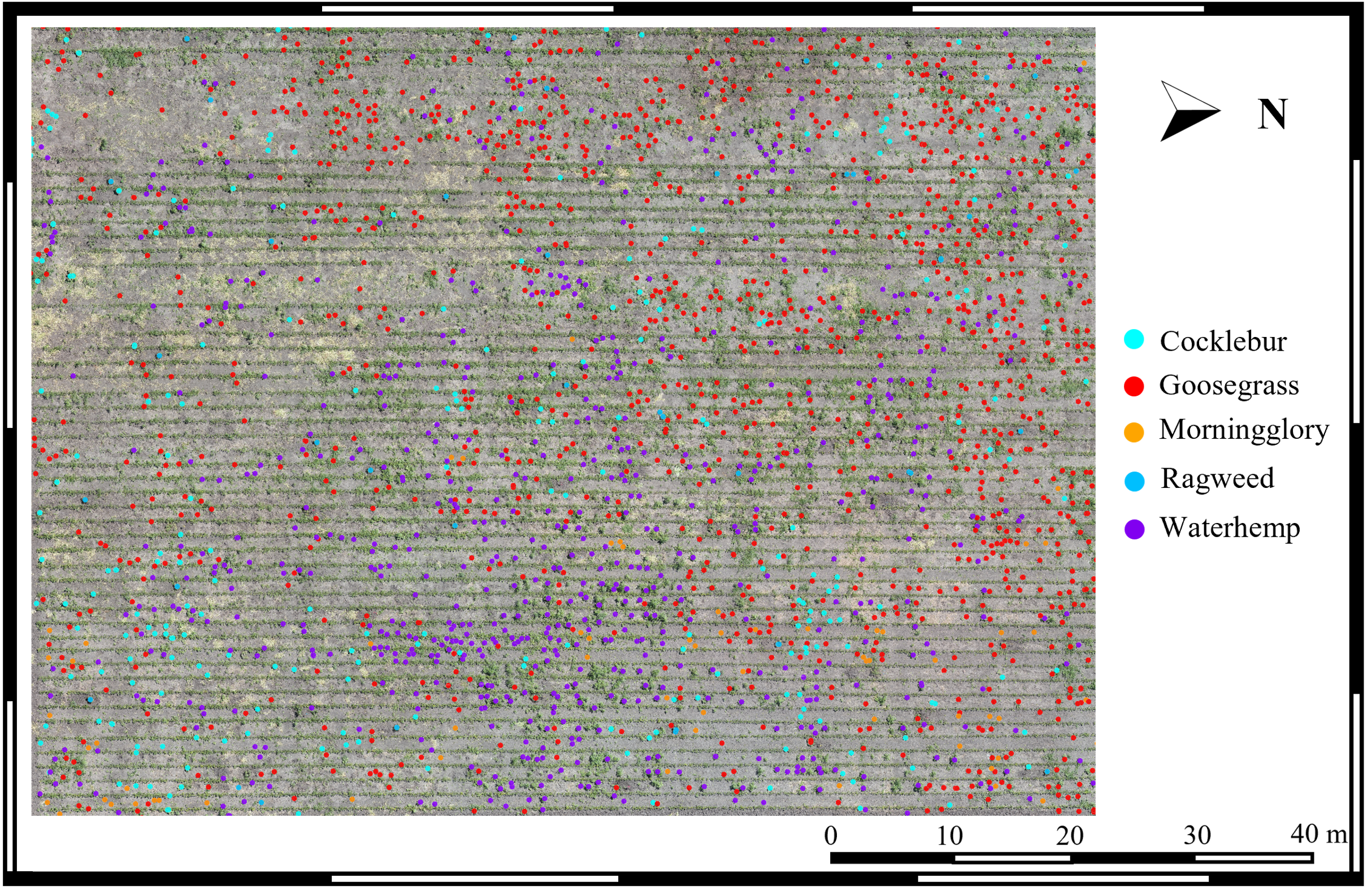} 
    \caption{Weed infestation map and weed species identification generated from detected weed coordinates.}
    \label{fig:weed_prescription}
\end{figure*}

Finally, as illustrated in Fig.~\ref{fig:weed_prescription}, a weed infestation map is generated by projecting detected bounding box centers into geographic space. The resulting spatial distribution exhibits clear clustering patterns rather than random dispersion. Specifically, Goosegrass (red) is predominantly concentrated on the right side of the field, while Waterhemp (purple) is clustered on the left. By generating prescription maps from such spatial insights, precision spraying systems can execute targeted, patch-based herbicide applications, thereby significantly reducing overall chemical inputs and improving resource efficiency.

\begin{table*}[htbp]
\centering
\caption{Detection accuracy comparison of two lightweight object detection models across different weed classes.}
\label{tab:weed_comparison}
\begin{tabular}{clcccccccccc}
\toprule
\multirow{2}{*}{Index} & \multirow{2}{*}{Weed Class} & \multicolumn{5}{c}{YOLOv12n} & \multicolumn{5}{c}{RT-DETRv1-R18} \\
\cmidrule(lr){3-7} \cmidrule(lr){8-12}
& & Precision & Recall & F1 & mAP$_{50}$ & mAP$_{50:95}$ & Precision & Recall & F1 & mAP$_{50}$ & mAP$_{50:95}$ \\
\midrule
1 & Cocklebur    & 86.49 & 84.21 & 85.33 & 85.71 & 47.38 & 88.89 & 63.16 & 73.85 & 80.49 & 46.45 \\
2 & Goosegrass   & 77.83 & 76.17 & 76.99 & 81.82 & 39.79 & 71.60 & 76.17 & 73.81 & 81.61 & 35.74 \\
3 & MorningGlory & 40.43 & 65.52 & 50.00 & 37.42 & 17.17 & 60.87 & 48.28 & 53.85 & 46.26 & 18.16 \\
4 & Ragweed      & 80.00 & 100.00 & 88.89 & 95.05 & 53.07 & 100.00 & 75.00 & 85.71 & 80.75 & 49.18 \\
5 & Waterhemp    & 75.97 & 69.01 & 72.32 & 75.42 & 37.76 & 60.71 & 71.83 & 65.81 & 67.55 & 32.03 \\
\bottomrule
\end{tabular}
\end{table*}

\subsection{Edge devices result analysis}
\begin{figure*}[htbp]
    \centering
    \includegraphics[width=0.95\textwidth]{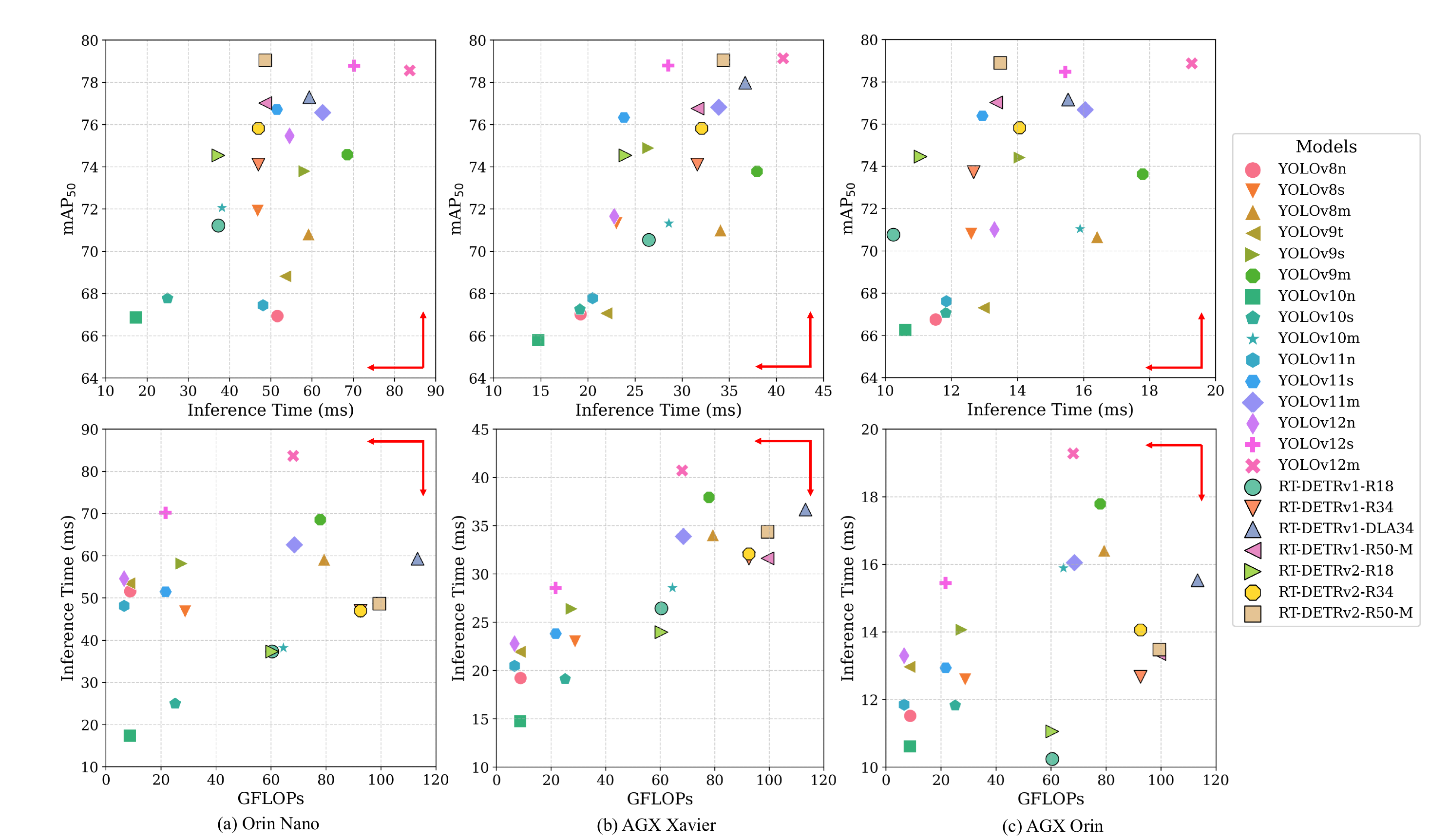} 
    \caption{Inference Speed and Detection Accuracy Comparison of Object Detection Models on Edge Devices.}
    
    \label{fig:edge_device}
\end{figure*}

Fig.~\ref{fig:edge_device} presents a comprehensive comparison of the YOLO and RT-DETR model families across three NVIDIA edge computing platforms:  Orin Nano, AGX Xavier, and AGX Orin. The top row illustrates the trade-off between detection accuracy ($\text{mAP}_{50}$) and inference time, while the bottom row shows the relationship between theoretical computational complexity (GFLOPs) and actual inference latency. 
Overall, a clear performance hierarchy is observed across the three devices. The Jetson AGX Orin delivers the fastest inference, with most models operating within 10–20 ms. The AGX Xavier demonstrates moderate performance, with inference times ranging from approximately 15–40 ms. In contrast, the Orin Nano exhibits the slowest performance, with latency spanning roughly 25–80 ms. These results highlight the significant impact of hardware capability on real-time deployment feasibility.

% comparison of the high accuracy models
From an accuracy perspective, YOLOv12m and RT-DETRv2-R50-M represent the top-performing models within the YOLO and Transformer-based families, respectively, both achieving $\text{mAP}_{50}$ values close to 80\% across all devices. However, despite comparable detection accuracy, YOLOv12m incurs substantially higher inference latency. In contrast, RT-DETRv2-R50-M demonstrates superior efficiency, achieving a more favorable balance between accuracy and speed.
For lightweight models, YOLOv10n and RT-DETRv2-R18 achieve the lowest inference latency across all platforms. However, this speed advantage comes at the cost of reduced detection accuracy, with $\text{mAP}_{50}$ values of approximately 66\% and 71\%, respectively. Considering practical deployment scenarios that require both efficiency and reliability, models such as YOLOv11s and RT-DETRv2-R50-M provide a more balanced trade-off between latency and detection performance.

Furthermore, analysis of the GFLOPs–latency relationship (bottom row) reveals that theoretical computational complexity does not directly translate to inference speed. Notably, YOLOv10 and RT-DETR models, which eliminate the need for NMS, achieve faster inference compared to traditional NMS-based YOLO variants, even when exhibiting similar or higher GFLOPs. This highlights the importance of end-to-end architectural design in optimizing real-world deployment performance.

\subsection{Discussion}\label{sec:discussions}

\subsubsection{Architectural impacts on small targets}
The performance discrepancy between YOLO and RT-DETR architectures, particularly on minority and small-scale classes, highlights the fundamental differences in their feature extraction paradigms. This divergence is visually captured in Table \ref{tab:weed_comparison}, where RT-DETR models demonstrate a clear advantage in detecting Morningglory, which typically occupies only a few tens of pixels. This superior performance on small targets can be attributed to the self-attention mechanism inherent in Transformer architectures, which effectively captures global contextual information and long-range dependencies. In contrast, conventional convolutional networks rely on localized receptive fields, which may limit their ability to model sparse or small-scale objects. These results suggest that while CNN-based models excel at extracting strong local textures, Transformer-based approaches are more robust when spatial information is limited or fragmented.
From an agronomic perspective, detection accuracies in the range observed here are likely sufficient to support early-season SSWM decisions, where timely identification of emerging weed patches is often more critical than perfect classification accuracy.

\subsubsection{Limitations and future directions}
This study has several limitations that warrant further investigation. First, although the primary goal was not to exhaustively evaluate all YOLO-based detectors, some recently developed models, such as YOLOv13 \citep{lei2025yolov13} and RT-DETRv3  \citep{wang2025rt}, were not included. Second, different precision formats and inference engines (e.g., FP32, NCNN) were not systematically evaluated. Future work will extend this benchmark to include these newer models and deployment configurations, providing a more comprehensive evaluation across diverse hardware environments.

\section{Conclusion}
This study presented a comprehensive benchmarking of the YOLO (v8-v12) and Transformer-based RT-DETR (v1-v2) model families for multi-species weed detection in crop fields under UAV-based imaging conditions. A systematic evaluation framework was established to compare detection accuracy, convergence behavior, and deployment efficiency across 37 models and multiple edge computing platforms.
The results demonstrated that RT-DETR models achieved a higher upper bound in detection accuracy, while the YOLO series maintained a competitive advantage in balancing accuracy and computational efficiency. Through class-specific analysis, this study revealed that Transformer-based architectures were more effective in handling small and spatially limited targets, particularly under challenging UAV-perspective conditions such as image degradation, occlusion, and scale variation.
In addition, this work provided an in-depth analysis of the trade-offs between theoretical complexity and real-world inference performance on edge devices. The findings showed that end-to-end architectures without NMS offered practical advantages in latency, highlighting the importance of considering deployment constraints beyond standard accuracy metrics. Based on these insights, YOLOv11s and RT-DETRv2-R50-M were identified as effective solutions for real-time applications, achieving a favorable balance between detection accuracy and inference speed.
Overall, this study contributed a practical and application-oriented benchmark that bridges algorithmic performance and deployment feasibility, offering actionable guidance for model selection in real-time, site-specific weed management systems.

\section*{Authorship Contribution}
\textbf{Linyuan Wang}: Conceptualization, Investigation, Software, Writing – Original Draft;
\textbf{Haibo Yao}: Conceptualization, Data preprocessing, Writing - Review ; 
\textbf{Te-Ming Tseng}: Conceptualization, Data Curation - Review \& Editing; 
\textbf{Kelvin Betitame}: Resources, Writing - Review \& Editing; 
\textbf{Xin Sun}: Resources, Writing - Review \& Editing; 
\textbf{Yanbo Huang}: Conceptualization, Data preprocessing, Writing - Review \& Editing; 
\textbf{Dong Chen}: Conceptualization, Investigation, Supervision, Writing - Review \& Editing.

\section*{Acknowledgement}

This publication is a contribution of the Mississippi Agricultural and Forestry Experiment Station (Starkville, MS). This research was funded by the Mississippi Soybean Promotion Board under the fund numbers (24-2025, 25-2026). Additionally, funding for this work was partly provided by USDA ARS in-house project No. 6064-21600-001-000-D and Cooperative Agreement No. 59-6064-5-002. The authors declare that the research was conducted in the absence of any commercial or financial relationships that could be construed as a potential conflict of interest. Mention of trade names or commercial products in this publication is solely for the purpose of providing specific information and does not imply recommendation or endorsement by the U.S. Department of Agriculture, Mississippi State University, and North Dakota State University.

\typeout{}
\bibliography{ref}
\end{document}